\documentclass[conference]{IEEEtran}
\usepackage{spconf,amsmath,graphicx}
\usepackage{booktabs}

\title{Transductive One-Shot Learning Meet Subspace Decomposition}
\name{Kyle Stein$^1$, Andrew A. Mahyari$^{1,2}$, Guillermo Francia, III$^3$, Eman El-Sheikh$^3$}
\address{$^1$ Department of Intelligent Systems and Robotics, University of West Florida, Pensacola, FL, USA\\
$^2$ Florida Institute For Human and Machine Cognition (IHMC), Pensacola, FL, USA \\
$^3$ Center for Cybersecurity, University of West Florida, Pensacola, FL, USA}

\begin{document}
%
\maketitle
\begin{abstract}
One-shot learning focuses on adapting pretrained models to recognize newly introduced and unseen classes based on a single labeled image. While variations of few-shot and zero-shot learning exist, one-shot learning remains a challenging yet crucial problem due to its ability to generalize knowledge to unseen classes from just one human-annotated image. In this paper, we introduce a transductive one-shot learning approach that employs subspace decomposition to utilize the information from labeled images in the support set and unlabeled images in the query set. These images are decomposed into a linear combination of latent variables representing primitives captured by smaller subspaces. By representing images in the query set as linear combinations of these latent primitives, we can propagate the label from a single image in the support set to query images that share similar combinations of primitives. Through a comprehensive quantitative analysis across various neural network feature extractors and datasets, we demonstrate that our approach can effectively generalize to novel classes from just one labeled image.

\end{abstract}
\section{Introduction}
\label{sec:intro}

One-shot learning (OSL) enables models to generalize and adapt to new tasks with minimal data~\cite{EnhancingUnlabelled, boudiaf2020information, Transductive_Tuning}. While traditional supervised models perform well with large labeled datasets, collecting and labeling such data is costly, especially in data-scarce fields. OSL allows models to recognize new objects from just one labeled example by leveraging prior knowledge from previously seen classes. This setup typically involves training on a single labeled support sample and evaluating on an unseen query set.

OSL techniques fall into two main categories: inductive and transductive. Inductive methods train a model solely on labeled support data, then apply it independently to predict on query samples~\cite{prototypical_networks, matching_networks, model_agnostic, RelationNet, DeepEMD, SimpleShot}. Transductive methods, by contrast, utilize the query set itself, finding feature similarities to labeled support samples to improve prediction accuracy, though they often require significant computational resources~\cite{LaplacianShot, ObliqueManifold, TPN, iLCT, EPNet}. State-of-the-art (SOA) transductive OSL techniques iteratively project query embeddings onto labeled supports for label propagation, yet they rarely exploit latent variables across classes. This can limit generalization on novel classes with similar compositional features.

In this paper, we introduce a data-driven approach based on subspace decomposition that achieves high accuracy while maintaining simplicity. Our method learns subspace bases and extracts latent variables from embeddings to enhance generalization on novel classes. The contributions of our paper include a method that simultaneously learns subspace bases for support and query sets, facilitating the extraction of latent compositional variables and leveraging insights from subspace decomposition and compositional zero-shot learning~\cite{huynh2020compositional, khan2023learning}. Inspired by prior work in subspace decomposition~\cite{lee1999learning, asteris2015orthogonal}, we also develop an unsupervised factorization technique that decomposes embeddings into subspaces representing distinct features, with support embeddings represented as linear combinations of these subspaces. \textbf{Unlike state-of-the-art methods that train models to directly classify images of new classes based on their feature vectors, our approach takes a novel turn by decomposing their feature vectors into class labels and subspace bases.}

\begin{figure*}[t]
    \centering
    \includegraphics[width=\textwidth, height=0.25\textheight, keepaspectratio]{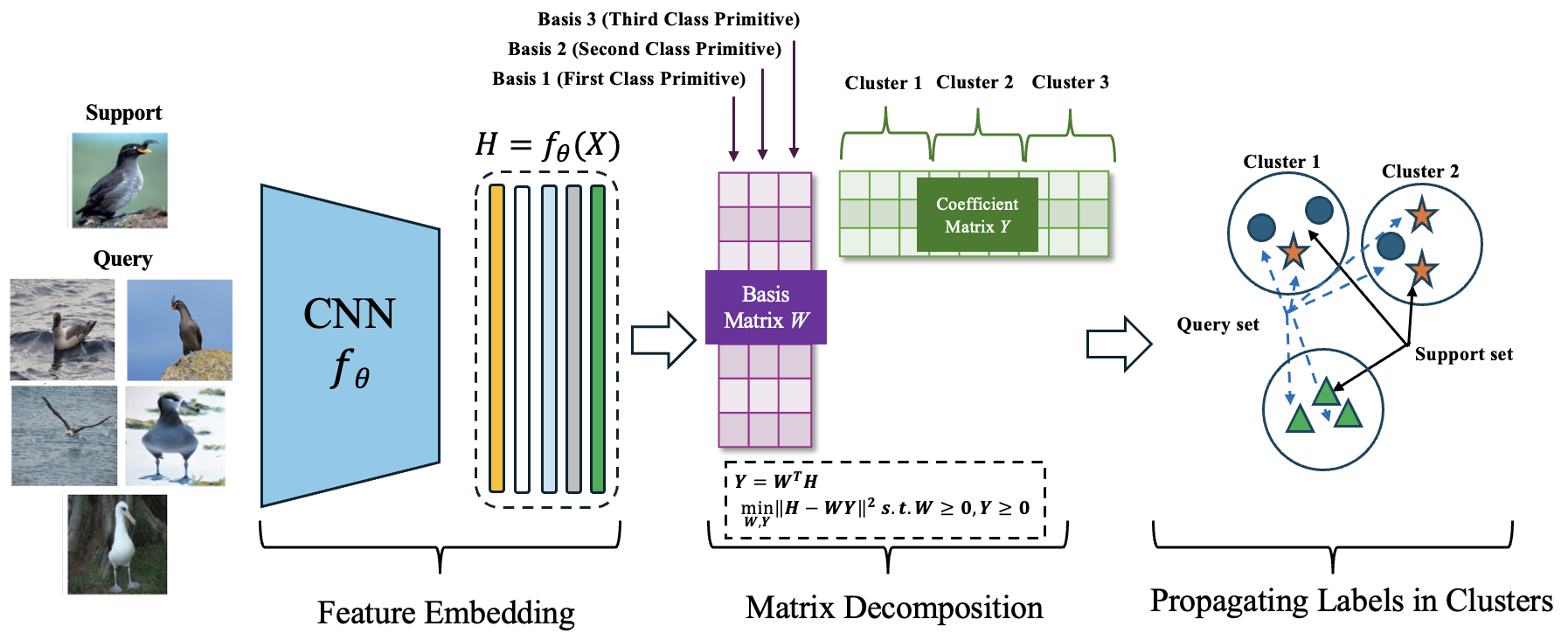} 
    \caption{\textit{Overall architecture of our approach for transductive one-shot learning. A pre-trained CNN extracts the features from the images, forming an embedding matrix. This matrix is then decomposed into a Basis Matrix and a Coefficient Matrix. The Basis Matrix contains fundamental class primitives, while the Coefficient Matrix encodes how these primitives combine to form image embeddings. The optimization process iteratively refines these matrices to minimize the reconstruction error. Finally, the Coefficient Matrix is used to propagate labels from the support set to the query set by classifying images with similar primitives.}}
    \label{fig:MainFig}
\vspace{-5mm}
\end{figure*}

\section{Related Work}
\label{sec:related_work}
Initial efforts in OSL aimed to reduce reliance on extensive annotated data, dividing approaches into metric-based and optimization-based methods. Metric-based methods train models to infer based on similarity measures in embedding spaces. Matching Networks~\cite{matching_networks} introduced cosine similarity for class embeddings, while Prototypical Networks~\cite{prototypical_networks} introduced class prototypes calculated as the mean embedding of support samples, assigning labels based on proximity in Euclidean space. 

Unlike metric-based approaches, optimization-based approaches in OSL focus on adapting model parameters to new tasks through fine-tuning with minimal updates. These methods aim to develop a model that can quickly adapt to new tasks with only a few gradient updates. Model-Agnostic Meta-Learning (MAML)~\cite{model_agnostic} popularized optimization approaches in OSL by training a model's parameters such that a small number of gradient updates will lead to quick adaptation and learning on a new task. Similar methods follow the same route by learning an optimal initialization that allows for efficient fine-tuning on new tasks with limited data. For example, \cite{temporal-convolutions} proposed a meta-learning approach that trains meta-learners on related tasks to generalize to new ones using temporal convolutions and soft attention, while \cite{Optimization-FSL} introduced an LSTM-based meta-learner designed to learn the specific optimization algorithm for training another neural network classifier. These approaches aim to minimize loss over a diverse set of tasks, training a base model that quickly generalizes and adapts to new scenarios. 

OSL can also be broken down into inductive and transductive approaches. Inductive approaches learn functions from support sets, independently predicting on query sets. In contrast, transductive methods access query data at inference, refining predictions. Laplacian Shot~\cite{LaplacianShot} employs Laplacian regularization for label consistency, while methods like~\cite{ObliqueManifold} and Transductive Propagation Network (TPN)~\cite{TPN} use joint feature spaces or graphs for label propagation.

\section{Preliminaries and Insights}

Let $S=\{ (x_i,y_i) \}_{i=1}^L$ represent $L$ labeled images of the support set, and let $Q=\{ (x_i) \}_{i=L+1}^{L+U}$ represent $U$ unlabeled images of the query set. In few-shot learning, we are given $K$ labeled images, $K$-shot, for $N$ classes,  N-way, known as the support set. We are also provided with a backbone feature extractor $f_\theta(\cdot)$ that maps the input raw images to the embedding ${\bf h}_i=f_\theta({\bf x}_i)$, where ${\bf h}_i \in {R}^{p \times 1}$. The goal of inductive few-shot learning is to learn a mapping or projection matrix ${\bf W} \in {R}^{p \times N}$ that maps the embedding to the correct labels ${\bf y}_i={\bf W}^T {\bf h}_i$, where ${\bf W}$ is learned from a small support set, and evaluated on the query set. The objective of the inductive few-shot learning is represented:

\vspace{-3mm}

\begin{equation}
    \min_{\bf W}{\sum_{i=1}^L{\mathcal{L}(y_i,{\bf W}^Th_i)}}
\end{equation}

In transductive few-shot learning, the relation between the features, \textit{i.e.} embedding, of the support and query sets is leveraged to generalize the projection matrix ${\bf W}$ to other unseen samples, and the cost function is given by: 

\begin{equation}
    \min_{\bf W}{\sum_{i=1}^L{\mathcal{L}(y_i,{\bf W}^Th_i)}+\sum_{i=1}^L{\sum_{j=1}^U{d(x_i,x_j)}}},
\end{equation}

\noindent where $d(\cdot,\cdot)$ is a similarity metric capturing the relationship among the samples of the support and query set. Our method builds on this by embedding both support and query samples into a shared feature space. We initialize the basis matrix ${\bf W}$ using the embedding of a labeled support sample and then construct a subspace that captures the relationship between embeddings of both support and query sets. Instead of relying on a predefined similarity metric, the relationships are infered by decomposing embeddings into latent components. The label from the support set is then propagated to the query set by comparing the coefficient vectors in the learned subspace.

\vspace{-2mm}
\section{Methodology}
\label{sec:methodology}
In this section, we examine the OSL problem from a subspace analysis perspective. Our approach aims to derive equations for learning subspaces that effectively represent the primitives of images in both support and query sets. By leveraging the subspace structure, we facilitate the classifying of images based on similar primitive combinations, enabling efficient label propagation in a transductive learning setting. While our method naturally groups similar features together in the subspace, we refer to this process as classifying, since the primary objective is to assign labels by aligning query images with the most relevant prototypes formed from a support sample.

To address the OSL problem, we derive equations for learning subspaces that best represent primitives of images across both support and query sets, and use these primitives for classification. Since the transductive approach leverages information from unlabeled samples in the query set, we combine the support and query sets into ${\bf X}=[{\bf x}_1, \ldots, {\bf x}_L, x_{L+1}, \ldots, {\bf x}_{L+U}]^T$. Our unsupervised method assumes that the labels for this set  ${\bf Y}$ are unknown, even though the label of one sample per class is known. We obtain the embeddings of the images in the set by passing them through a backbone feature extractor, resulting in ${\bf H}=f_\theta({\bf X})=[{\bf h}_1,\ldots,{\bf h}_{L+U}]^T$. 

Similar to prior works \cite{EASE+Kmeans}, we assume that the labels of the support and query sets can be predicted by a linear projection of the embeddings onto the output manifold, which we capture with ${\bf Y}={\bf W}^T{\bf H}$. Since ${\bf Y}$ and ${\bf W}$ are unknown in this equation, we rearrange the subspace projection equation as ${\bf H}={\bf W}{\bf Y}$, where ${\bf W}$ is orthonormal. The label matrix ${\bf Y}$ is sparse; thus, this equation is interpreted as a sparse representation of the embeddings ${\bf H}$, where the columns of ${\bf W}$ are the basis of a subspace, and ${\bf Y}$ represents the coefficient vectors for this basis. The matrix ${\bf H}$ is thus projected onto the subspace defined by the columns of ${\bf W}$. Ideally, each embedding vector ${\bf h}_i$ is represented by one basis (i.e., one column) of the basis matrix (i.e., the projection matrix) ${\bf W}$. In the special case of OSL, each column of ${\bf W}$ could be equivalent to the embedding vector of the support sample, ${\bf h}_i={\bf w}_i$. This simplifies to the average of the embeddings of the samples per class in the supporting set, resulting in a Protoypical network \cite{prototypical_networks}.

The embedding matrix derived from input images through the backbone feature extractor is the result of {\sffamily{ReLU}} operations, thus ensuring that the embeddings are always non-negative, ${\bf H} \geq 0$. This is consistent with common practices in deep learning architectures, where {\sffamily{ReLU}} activation functions are incorporated to introduce non-linearity while avoiding the vanishing gradient problem \cite{WRN-28-10, ResNet-12}. Similarly, the coefficient matrix, which represents the labels or the distribution over the classes, is also non-negative, and each row (${\bf y}_i$) represents the output distribution, therefore we have two additional conditions: ${\bf Y} \geq 0$ and $\sum_j{softmax({\bf y}_i)_j}=1$. The second constraint, \textit{i.e.,} $\sum_j{softmax({\bf y}_i)_j}=1$, ensures the the estimated coefficients sum up to one after passing through a \emph{softmax} operator, representing the categorical distribution over classes. Incorporating these constraints into the linear relationship between the embeddings and the output labels shapes our primary objective function:


\vspace{-5mm}
\begin{equation}\label{eq:subspace1}
\begin{array}{c}
    \min_{{\bf Y},{\bf W}}{\|{\bf H}-{\bf W}{\bf Y}\|_F^2} \hspace{2mm} \\
    s.t. \hspace{2mm} {\bf W} \geq 0, {\bf Y} \geq 0, \sum_j{softmax({\bf y}_i)_j}=1
\end{array}
\end{equation}

\vspace{-2mm}
Eq.~\ref{eq:subspace1} depicts a problem of simultaneous sparse representation and dictionary learning, where ${\bf W}$ functions as an unknown dictionary and ${\bf Y}$ as the sparse representation of the embeddings relative to this dictionary ${\bf W}$. Although various dictionary learning methods could be employed to determine ${\bf W}$ and ${\bf Y}$, we opt for matrix decomposition to address this optimization challenge.

\begin{table*}[h!]
    \renewcommand{\arraystretch}{1.1}
    \centering
    \begin{tabular}{l|c|c|c|c} 
        \toprule
        \textbf{Method} & \textbf{Setting} & \textbf{Backbone} & \textbf{mini-ImageNet (1-shot)} & \textbf{tiered-ImageNet (1-shot)} \\
        \midrule
        MAML \cite{model_agnostic} & Inductive & ResNet-18 & 49.61 $\pm$ 0.92 & - \\
        RelationNet \cite{RelationNet} & Inductive & ResNet-18 & 52.48 $\pm$ 0.86 & - \\
        MatchingNet \cite{matching_networks} & Inductive & ResNet-18 & 52.91 $\pm$ 0.88 & - \\
        ProtoNet \cite{prototypical_networks} & Inductive & ResNet-18 & 54.16 $\pm$ 0.82 & - \\
        DeepEMD \cite{DeepEMD} & Inductive & ResNet-18 & 65.91 $\pm$ 0.82 & - \\
        TPN \cite{TPN} & Transductive & ResNet-12 & 55.51 $\pm$ 0.86 & 59.91 $\pm$ 0.94 \\
        Transductive Tuning \cite{Transductive_Tuning} & Transductive & ResNet-12 & 62.35 $\pm$ 0.66 & - \\
        DSN-MR \cite{DSN-MR} & Transductive & ResNet-12 & 64.60 $\pm$ 0.72 & 67.39 $\pm$ 0.82 \\
        CAN-T \cite{CAN-T} & Transductive & ResNet-12 & 67.19 $\pm$ 0.55 & 73.21 $\pm$ 0.58 \\
        EASE \cite{EASE+Kmeans} & Transductive & ResNet-12 & 57.00 $\pm$ 0.26 & 69.74 $\pm$ 0.31 \\
        \midrule
        \textbf{Proposed} & Transductive & ResNet-12 & \textbf{67.55 $\pm$ 0.24} & \textbf{81.06 $\pm$ 0.49} \\
        \midrule
        ProtoNet \cite{prototypical_networks} & Inductive & WRN-28-10 & 62.60 $\pm$ 0.20 & - \\
        MatchingNet \cite{matching_networks} & Inductive & WRN-28-10 & 64.03 $\pm$ 0.20 & - \\
        SimpleShot \cite{SimpleShot} & Inductive & WRN-28-10 & 65.87 $\pm$ 0.20 & 70.90 $\pm$ 0.22 \\
        Transductive Tuning \cite{Transductive_Tuning} & Transductive & WRN-28-10 & 65.73 $\pm$ 0.68 & 73.34 $\pm$ 0.71 \\
        TIM \cite{boudiaf2020information} & Transductive & WRN-28-10 & 77.80 & 82.10 \\
        EPNet \cite{EPNet} & Transductive & WRN-28-10 & 70.74 $\pm$ 0.85 & 78.50 $\pm$ 0.91 \\
        LaplacianShot \cite{LaplacianShot} & Transductive & WRN-28-10 & 74.86 $\pm$ 0.19 & 80.18 $\pm$ 0.21 \\
        Oblique Manifold \cite{ObliqueManifold} & Transductive & WRN-28-10 & \textbf{80.64 $\pm$ 0.34} & \textbf{85.22 $\pm$ 0.34} \\
        EASE \cite{EASE+Kmeans} & Transductive & WRN-28-10 & 67.42 $\pm$ 0.27 & 75.87 $\pm$ 0.29 \\
        \midrule
        \textbf{Proposed} & Transductive & WRN-28-10 & 76.96 $\pm$ 0.60 & 84.55 $\pm$ 0.33 \\
        \bottomrule
    \end{tabular}
    \caption{Test accuracy vs. the state-of-the-art (1-shot classification) on mini-ImageNet and tiered-ImageNet. }
    \label{tab:combined_cases}
    \vspace{-5mm}
\end{table*}

In Eq.~\ref{eq:subspace1}, the matrix decomposition approach breaks down the embedding matrix ${\bf H}$ into two components: the unknown projection matrix ${\bf W}$ and the coefficient matrix ${\bf Y}$. Each embedding vector ${\bf h_i}$ in ${\bf H}$ is approximated as a combination of the basis vectors in ${\bf W}$, weighted by the coefficients in the corresponding column in ${\bf Y}$. Each column of ${\bf W}$ serves as a latent feature vector, encapsulating a primitive within the embedding matrix \cite{wang2012nonnegative}. Given that the number of columns in ${\bf W}$ is significantly fewer than the dimension of the embedding vectors, this decomposition method characterizes each class primarily by one dominant primitive. Consequently, these primitives contain the main distinguishing feature of each class, allowing images from both the support and query sets to be classified based on their dominant primitive \cite{wang2012nonnegative}. 

The coefficient matrix ${\bf Y}$ represents the relationship between the basis vectors in ${\bf W}$ to the representation of the embedding vector ${\bf h_i}$. Each column of ${\bf Y}$ represents a coefficient vector which captures the combination of primitives that are shared among images. These coefficient vectors indicate how different embeddings are represented within the learned subspace ${\bf W}$. The similarity between these coefficient vectors allows us to classify the query sample features to those of a single labeled support sample, leveraging the shared subspace for label propagation. Eq.~\ref{eq:subspace1} is convex with respect to either ${\bf W}$ or ${\bf Y}$. To solve it, we employ gradient descent, iteratively estimating ${\bf W}$ and ${\bf Y}$. It is worth noting that this optimization involves only a few samples, allowing us to perform the calculations in one step without requiring stochastic gradient descent. Furthermore, overfitting is not a concern since we are not learning parameters but instead decomposing the matrix ${\bf H}$ into the product of two matrices, akin to matrix decomposition techniques. However, because the closed-form solution for Equation \ref{eq:subspace1} cannot be derived, we rely on gradient descent to approximate the solution.



To initialize ${\bf Y}$, known labels from the support set are one-hot encoded. During optimization, the alternating update of ${\bf W}$ and ${\bf Y}$ iteratively adjusts the representation of both the projection matrix and coefficient representation, minimizing the reconstruction error $\|{\bf H}-{\bf W}{\bf Y}\|_F^2$. By minimizing the reconstruction error, the model identifies the most discriminative features in the data, encouraging similar samples in the latent space to classify based on shared primitives. As optimization converges, ${\bf Y}$ captures the label distribution for both support and query samples. The predicted label for each sample is determined by the position of the maximum value in its corresponding column of ${\bf Y}$. 

Up to this point, we have approached the OSL task as an unsupervised task, aiming to classify query images based on similar primitives with a support image. After establishing the classes, we propagate the known label from a single support image to all query images within the same class. 

\section{Experiments and Results}
\label{sec:experiment}

\begin{table*}[t!]
\centering
\small
\begin{tabular}{l|c|c|c}
\toprule
\textbf{Model} & \textbf{Setting} & \textbf{mini-ImageNet Accuracy (\%)} & \textbf{tiered-ImageNet Accuracy (\%)} \\
\midrule
MAML~\cite{model_agnostic} & Inductive & 31.27 $\pm$ 1.15 & 34.44 $\pm$ 1.19 \\
MAML+Transduction~\cite{model_agnostic} & Transductive & 31.83 $\pm$ 0.45 & 34.78 $\pm$ 1.18 \\
ProtoNet~\cite{prototypical_networks} & Inductive & 32.88 $\pm$ 0.47 & 37.35 $\pm$ 0.56 \\
RelationNet~\cite{RelationNet} & Inductive & 34.86 $\pm$ 0.48 & 38.62 $\pm$ 0.57 \\
TPN~\cite{TPN} & Transductive & 36.62 $\pm$ 0.50 & 40.93 $\pm$ 0.61 \\
Simple CNAPS~\cite{bateni2020improved} & Transductive & 37.10 $\pm$ 0.50 & 48.10 $\pm$ 0.70 \\
Transductive CNAPS~\cite{EnhancingUnlabelled} & Transductive & 42.80 $\pm$ 0.70 & 54.60 $\pm$ 0.80 \\
\midrule
\textbf{Proposed Method} & Transductive & \textbf{47.03 $\pm$ 0.18} & \textbf{63.26 $\pm$ 0.19} \\
\bottomrule
\end{tabular}
\caption{1-shot 10-way accuracy results with 10 query samples for various models on mini-ImageNet and tiered-ImageNet.}
\label{tab:10way_1shot_results}
\vspace{-5mm}
\end{table*}

\subsection{Datasets and Benchmarks}
Multiple datasets were chosen to validate our method, notably: mini-ImageNet~\cite{matching_networks} and tiered-ImageNet~\cite{tiered}. These datasets are commonly inferred upon in the OSL community due to their complexity and diversity, which make them ideal for evaluating the generalization capabilities of these models. \textbf{MiniImageNet} consists of 60,000 colour images with 100 classes, each having 600 examples. \textbf{Tiered-ImageNet} represents a larger subset of classes from ILSVRC-12 than mini-Imagenet, with 608 classes. Not only do more classes exist, but this dataset also provides a more structured hierarchy, which ensures that all of the training classes are sufficiently distinct from the testing classes. 

Our method is compared to other inductive and transductive SOA results present in the literature: MAML ~\cite{model_agnostic}, RelationNet ~\cite{RelationNet}, MatchingNet ~\cite{matching_networks}, ProtoNet ~\cite{prototypical_networks}, DeepEMD ~\cite{DeepEMD}, TPN ~\cite{TPN}, Transductive Tuning ~\cite{Transductive_Tuning}, DSN-MR ~\cite{DSN-MR}, CAN-T ~\cite{CAN-T}, EASE ~\cite{EASE+Kmeans}, SimpleShot ~\cite{SimpleShot}, TIM ~\cite{boudiaf2020information}, Boosting \cite{gidaris2019boosting}, EPNet ~\cite{EPNet}, LaplacianShot ~\cite{LaplacianShot}, and Oblique Manifold ~\cite{ObliqueManifold}.

\subsection{Experimental Setup}
Episodic training is a widely utilized technique in few-shot learning, particularly in OSL scenarios. This method mimics the test conditions where the model is exposed to a limited number of labeled samples $S$ and is expected to generalize to unlabeled examples from $Q$. Each training episode involves a $N$-way, $K$-shot task. This task is set up by selecting a subset of $N$ classes from the training set. From each class in this subset, $K$ samples are randomly chosen to create the labeled support set $S$. Additional random samples from these $N$ classes are selected to form the query set $Q$. During each episode, the feature extractor $f_\theta(\cdot)$ processes both $S$ and $Q$ to generate embeddings. The embeddings from $S$ are utilized to train ${\bf W}$, which is then applied to the embeddings of $Q$ for label prediction. The accuracy of these predictions is assessed by comparing them with the true labels of the query set. The discrepancy, measured as loss, is used to refine the parameters $\theta$ of $f_\theta(\cdot)$.

Our initial experiment, in Table \ref{tab:combined_cases}, involves 10,000 randomly generated episodes, each following a 5-way, 1-shot format with 15 query samples per episode. To conduct further analysis, we extend the experimental setup in Table \ref{tab:10way_1shot_results} to a more challenging 10-way, 1-shot scenario, while keeping the number of episodes the same and reducing the number of query samples to 10. We conduct each experiment 5 times, calculating the mean between the experiments and 95\% confidence intervals for consistency. For our analysis, we employ pre-trained feature extractors: ResNet-12 ~\cite{ResNet-12} and WRN-28-10 ~\cite{WRN-28-10} as $f_\theta(\cdot)$ to extract embeddings from input images. 

\subsection{Results}
The experimental results on mini-ImageNet and tiered-ImageNet are shown in Table \ref{tab:combined_cases}. We show SOA performance for OSL across both datasets. We can observe that the proposed method outperforms the SOA methods for image classification on the tiered-ImageNet when using extracted features from ResNet. We improve accuracy by nearly 8\% over the nearest method using one labeled support sample. On mini-ImageNet, with features extracted using ResNet, we also obtain the highest classification accuracy. When employing the features extracted from WRN-28-10, we can see overall improved performance of our method when compared to ResNet. However, WRN-28-10’s larger feature space ($p$ = 640) expands the search space for our non-convex subspace decomposition, making the alternating updates more prone to settle in suboptimal stationary points rather than find the global optimum. This complicates the task of identifying appropriate primitives (columns of ${\bf W}$) within a more expansive and complex search space. The performance variability demonstrates the challenges of different architectures and suggests that all methods have specific strengths and limitations depending on the experimental setup.


To test the robustness of our model across different scenarios, we increased the number of classes during inference from 5 to 10, while reducing the number of query samples to 10, following the approach presented in \cite{EnhancingUnlabelled}. Table \ref{tab:10way_1shot_results} displays our method's results using ResNet-12 in comparison with other SOA OSL methods. We observe that our method outperforms previous methods in the 10-way classification scenario. Specifically, our model improves accuracy by over 4\% on mini-ImageNet and achieves an impressive increase of more than 9\% on tiered-ImageNet. To the best of our knowledge, these are SOA results for 10-way accuracy on both mini-ImageNet and tiered-ImageNet. This performance increase shows the robustness of our method, even when tasked with handling a more complex classification task. 


The efficient performance of our model can be attributed to subspace decomposition, which provides a more refined representation of data in the latent space. This method enables us to effectively utilize the information from a single labeled support sample to extend the labels to the query samples within the same subspace. This process ensures that the embedding vectors projected onto the subspace establish a clearer connection between the support and query samples. This achievement is facilitated by the concurrent learning of the basis and coefficient matrices. Additionally, the cost function plays a crucial role in enhancing the stability and overall performance of the model. These results confirm that our subspace decomposition method, regardless of the feature extractor, enables efficient label propagation and classification in challenging one-shot scenarios.

\section{Conclusion}
\label{sec:conclusion}

In this paper, we introduced a novel transductive OSL approach that identifies primitives of images by decomposing the embeddings of images from both support and query sets into representative subspaces. While our method demonstrates high accuracy, further extensive research is necessary to explore this data-driven approach, particularly to understand the impact of hidden factors and their connections to both seen and unseen classes. The empirical study revealed that the variability in performance demonstrates the inherent challenges posed by different architectures, suggesting that each method has specific strengths and limitations influenced by both the experimental setup and the nature of the datasets. Future efforts in this area will aim to expand this data-driven subspace decomposition methodology to zero-shot learning, linking attribute vectors to the primitives extracted through subspace factorization techniques.

\bibliographystyle{IEEEbib}
\bibliography{strings,refs}

\end{document}